\documentclass{article}

\usepackage{natbib}
\usepackage[preprint]{neurips_2023}
\usepackage[utf8]{inputenc} % allow utf-8 input
\usepackage[T1]{fontenc}    % use 8-bit T1 fonts
\usepackage{url}            % simple URL typesetting
\usepackage{booktabs}       % professional-quality tables
\usepackage{amsfonts}       % blackboard math symbols
\usepackage{nicefrac}       % compact symbols for 1/2, etc.
\usepackage{microtype}      % microtypography
\usepackage{colortbl}
\usepackage[table]{xcolor}
\definecolor{darkgreen}{HTML}{3cb44b}
\definecolor{darkred}{HTML}{ff7256}
\definecolor{xdcolor}{HTML}{19b5af}
\usepackage{multirow}
\usepackage[hidelinks]{hyperref}
\usepackage{graphicx}
\usepackage{xspace}
\usepackage{geometry}
\usepackage{changepage}
\usepackage{arydshln}

\newcommand{\llm}{Xmodel-LM\xspace}

\title{\llm Technical Report}
\author{
Wang Yichuan
\hspace{0.8em}Liu Yang
\hspace{0.8em}Yan Yu
\hspace{0.8em}Wang Qun
\hspace{0.8em}Huang Xucheng
\hspace{0.8em}Jiang Ling \\ \\
XiaoduoAI \\
\texttt{\{wangyichuan,liuyangfoam,yanyu,wangqun\}@xiaoduotech.com}
}
\date{}

\begin{document}

\maketitle

\begin{abstract}

We introduce \llm, a compact and efficient 1.1B language model pretrained on around 2 trillion tokens. Trained on our self-built dataset (Xdata), which balances Chinese and English corpora based on downstream task optimization, \llm exhibits remarkable performance despite its smaller size. It notably surpasses existing open-source language models of similar scale. Our model checkpoints and code are publicly accessible on GitHub at \url{https://github.com/XiaoduoAILab/XmodelLM}.

\end{abstract}

\section{Introduction}

Large language models (LLMs) have demonstrated remarkable performance across various natural language tasks, often requiring minimal examples of natural language instructions \citep{brown2020language}, thereby alleviating the necessity for extensive feature engineering.

At the heart of this progress lies language modeling, typically formulated as an unsupervised distribution estimate from sequences of symbols $(s_1, ..., s_{n-1})$ within a variable-length sequence $x$. Leveraging the natural sequential order of languages, the joint probability over symbols is commonly factorized into a product of conditional probabilities:

$$
    p(x) = 
    \prod_{i=1}^{n} p ( s_n | s_1, ..., s_{n-1})
$$

Recent advancements in natural language processing (NLP) have largely stemmed from the scaling up of language model sizes, driven by the observed functional relationship between model performance and size \citep{henighan2020scaling}. However, the accompanying rise in operational costs poses a significant hurdle to their widespread adoption.

In this paper, we introduce \llm, a compact vision-language assistant powered by a relatively small language model. Remarkably, \llm achieves performance comparable to state-of-the-art models of similar scale across numerous LLM benchmark tests, showcasing its potential for a wide array of practical tasks.

\section{Pretraining}
This chapter details the pretraining process of \llm. First, we introduce the sources and composition of our corpus, as well as our preprocessing methods. Second, we describe the construction of our customized tokenizer. Finally, we detail the model architecture and training parameter configurations.

\subsection{Training Data}

\textbf{Data sourcing}: In the process of constructing the training corpus and allocating weights, our primary objective is to ensure the quality and diversity of the training data. Our original dataset primarily consists of aggregated training data from other LLMs, such as Redpajama \citep{together2023redpajama}, subsets of the Pile \citep{gao2020pile} and StarCoder \citep{li2023starcoder}. To address deficiencies in the distribution of book and mathematical data within the training data distribution, we have also incorporated FanFics\footnote{https://huggingface.co/datasets/marianna13/fanfics} and OpenWebMath \citep{paster2023openwebmath}. Additionally, we have added the Chinese data source PTD \citep{wang2024telechat} and WanJuan \citep{he2023wanjuan} to imbue our model with a certain level of proficiency in Chinese.

\textbf{Data processing}: We are committed to ensure the quality of the data and reducing its redundancy. We first employ heuristic methods such as paragraph length and punctuation ratio for initial filtering. Subsequently, we utilize a 5-gram Kneser-Ney model based on KenLM Library \citep{heafield-2011-kenlm} to compute text perplexity for further quality filtering. In the next stage, we employe a locality-sensitive hashing method based on SimHash to deduplicate the training data. To balance deduplication quality and efficiency, we implement a bucketing strategy on the entire dataset, enabling the deduplication process to scale efficiently across large datasets. Finally, we tokenize all datasets using our custom-trained tokenizer, and designed different sampling weights for the datasets based on their characteristics, as shown in Table~\ref{tab: data_info}.

\begin{table}[ht]
  \centering
  \setlength{\tabcolsep}{2pt}
  \begin{tabular}{@{}lcccccc@{}}
    \noalign{\hrule height 1.2pt}
    Data Source & Dataset & Num Tokens & Sampling Weight & Epochs & Category & Language \\ \midrule

    \multirow{7}{*}{Redpajama}& Arxiv & 31,336,679,261 & 0.0160 & 1 & Academic & English \\
    & Book & 29,633,248,538 & 0.0300 & 1 & Book & English \\
    & C4 & 192,696,661,887 & 0.1000 & 1 & Web & English \\
    & Common Crawl & 1,251,868,330,446 & 0.5600 & 0.88 & Web & English \\
    & Github & 59,063,773,003 & 0.0150 & 0.5 & Code & English \\
    & Stackexchange & 22,728,030,774 & 0.0174 & 1.5 & Social & English \\
    & Wikipedia & 34,312,919,854 & 0.0520 & 3 & Academic & English \\ \midrule

    \multirow{2}{*}{Pile} & BookCorpus & 562,392,085 & 0.0006 & 2 & Book & English \\
    & PubMed & 17,698,877,602 & 0.0100 & 1 & Academic & English \\ \midrule

    AMPS & AMPS & 269,936,326 & 0.0003 & 2 & Math & English \\ \midrule
    FanFics & FanFics  & 1,803,437,344 & 0.0020 & 2 & Book & English \\ \midrule
    OpenWebMath & OpenWebMath & 7,150,335,312 & 0.0080 & 2 & Math & English \\ \midrule
    StarCoder & StarCoder & 306,812,862,958 & 0.0536 & 0.3 & Code & English \\ \midrule

    \multirow{4}{*}{WanJuan} & Law & 9,080,387,832 & 0.0100 & 2 & Academic & Chinese \\
    & News & 5,175,531,875 & 0.0050 & 2 & Academic & Chinese \\
    & Patent & 4,559,904,057 & 0.0050 & 2 & Academic & Chinese \\
    & Webtext & 126,429,462,230 & 0.0644 & 1 & Web & Chinese \\ \midrule

    PTD & PTD & 165,879,069,486 & 0.0507 & 0.6 & Web & Chinese \\
    
    \noalign{\hrule height 1.2pt}
    
  \end{tabular}
  \newline
  \caption{Detailed composition of the training set.}
  \label{tab: data_info}
\end{table}

\subsection{Tokenizer}
We employ the unigram algorithm \citep{kudo2018subword} for data tokenization, utilizing the implementation provided by Sentence-Piece \citep{kudo2018sentencepiece}. In contrast to the extensive vocabularies used in prevailing open-source models, our tokenizer is trained on a mixed corpus of Chinese and English, with a vocabulary size of only 32,000. The comparison of the \llm tokenizer with other tokenizers is shown in the Table~\ref{tab: tokenizer}. Despite its small size, our tokenizer demonstrates impressive compression rates on test data.

\begin{table}[!ht]
    \centering
    \setlength{\tabcolsep}{20pt}
    \begin{tabular}{lcc}
		\noalign{\hrule height 1.2pt}
		Tokenizer & Vocab Size & Compression Rate $\downarrow$ \\ 
        \midrule
		LLaMA 2     & 32,000      & 0.7524          \\
		InternLM 2      & 103,168     & 0.4124            \\
		Baichuan 2  & 125,696      & 0.4103           \\
		\llm  & 32,000     & {\bf 0.3841}          \\ 
        \noalign{\hrule height 1.2pt}
	\end{tabular}
 \newline
	\caption{Comparison of vocabulary size and text compression rate of \llm's tokenizer with other models. Lower values indicate better compression.}\label{table.tokenzier}
 \label{tab: tokenizer}
\end{table}

The \llm tokenizer is trained using a subset of the \llm pre-training corpus, without any applied text normalization. To improve the encoding of numeric data, numbers are split into individual digits. Character coverage is set to 0.9999, with rare characters being represented by UTF-8 bytes. Additionally, we set the maximum token length to 16 to accommodate Chinese phrases.

\subsection{Model architecture}

We adopt a similar model architecture to LLama 2 \citep{touvron2023llama} with the following details:

\begin{table}[ht]
  \centering
  \setlength{\tabcolsep}{6pt}
  \begin{tabular}{@{}cccccc@{}}
    \toprule
    Hidden size & Intermediate size & Attention heads &  KV heads & Layers & Context Len\\
    \midrule
    2048 & 5632 & 32 & 4 & 24 & 4096\\
    \bottomrule
  \end{tabular}
  \newline
  \caption{Detailed settings of \llm.}
  \label{tab: LLM_setting}
\end{table}

\noindent\textbf{Rotary Positional Embedding.} We integrate rotary positional embeddings (RoPE) \citep{su2023roformer} at each layer of the network.

\noindent\textbf{RMSNorm.} To enhance training stability, we utilize the RMSNorm \citep{Zhang2019RMSNorm} function to normalize the input of each transformer sub-layer, without normalizing the output. Linear layers do not incorporate bias, and word embeddings are not tied.

\noindent\textbf{SwiGLU.} We replace the conventional ReLU non-linearity with the SwiGLU \citep{shazeer2020glu} activation function to optimize performance.

\noindent\textbf{Grouped-query attention.} For efficient training and inference, we employ grouped-query attention (GQA) \citep{ainslie2023gqa}, featuring 32 attention heads and 4 KV heads.

\subsection{Training}

Training is conducted on a single node equipped with 7$\times$H800 GPUs. To enhance training efficiency and boost Model FLOPS Utilization (MFU), we employ Distributed Data Parallel (DDP) and FlashAttention-V2.

We utilize the cumulative gradient updating method with a mini-batch size of 4 and a gradient accumulation step of 30 per GPU, resulting in a global batch size of 840 with a sequence length of 4096. This setup yields a total token count of 3,440,640 per iteration. Training spans 600,000 iterations, accumulating to a total token count of 2,064,384,000,000.

For optimization, we employ the AdamW optimizer with a maximum learning rate of 6e-4. The learning rate undergoes linear increase from zero to the maximum over the first 2000 updates, followed by annealing to 6e-5 using a cosine schedule. The batch size is set to around 3.5M tokens, with weight decay assigned as 0.1. Additionally, we apply a gradient clipping threshold of 1.0 to regulate the gradient value.

\begin{figure}[ht]
\centering
\includegraphics[width=\linewidth]{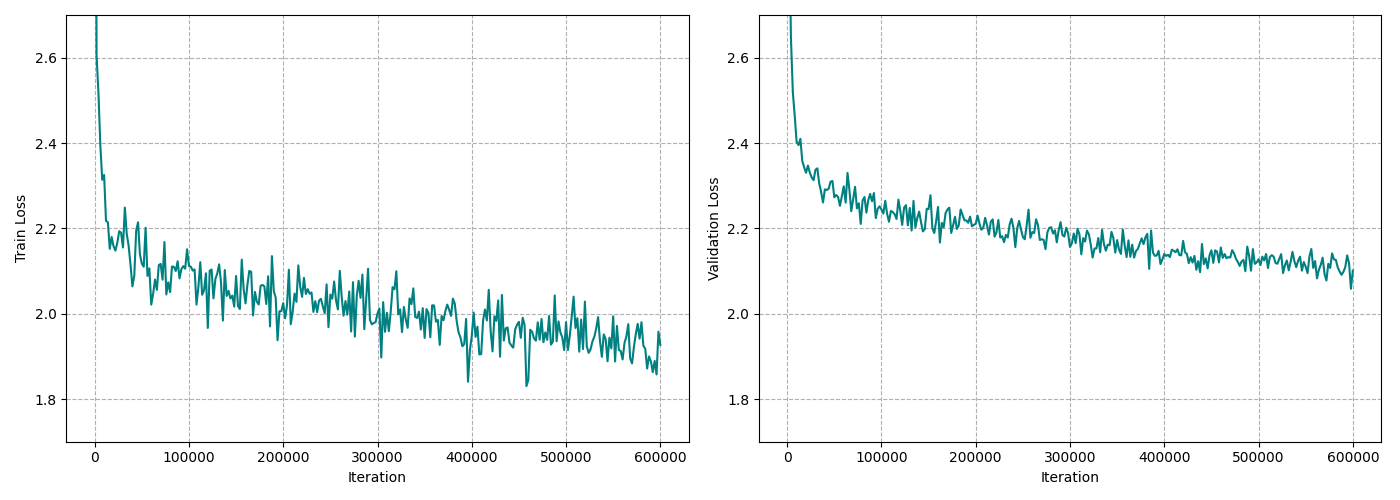}
\caption{The trend of training and validation loss during pretraining.}
\label{fig:loss}
\end{figure}

Refer to the training log in the Figure~\ref{fig:loss}, which includes trend graphs showing the training and validation losses as the training token count increases. We use OpenWebText \citep{Gokaslan2019OpenWeb}, which is not included in the training set, as the validation set to calculate the validation loss.

\section{Results}
\label{sec:results}

\noindent\textbf{Baseline models} For the sake of comparability, we chose several recently popular decoder-only architecture models with a scale of around 1 billion parameters for comparison. Specifically, we compare \llm with OPT \citep{zhang2022opt}, Pythia \citep{biderman2023pythia}, TinyLLaMA \citep{zhang2024tinyllama}, MobileLLaMA \citep{thawakar2024mobillama}, H2O-danube \citep{singer2024h2odanube18b}, InternLM2 \citep{cai2024internlm2} and Qwen1.5 \citep{qwen}. % Phi-1.5 \citep{li2023textbooks} % StableLM-2 \citep{bellagente2024stable}, MPT \citep{MosaicML2023Introducing} Gemma \citep{gemmateam2024gemma}

\noindent\textbf{Commonsense reasoning tasks} For evaluating our models we use the Language Model Evaluation Harness \citep{eval-harness}. Specifically, our evaluation comprised: ARC-Challenge \citep{clark2018think}, ARC-Easy \citep{clark2018think}, Boolq \citep{clark2019boolq}, HellaSwag \citep{zellers2019hellaswag}, OpenBookQA \citep{mihaylov2018suit}, PiQA \citep{Bisk2019PIQARA},  SciQ \citep{welbl2017crowdsourcing}, TriviaQA \citep{joshi2017triviaqa}, Winogrande \citep{Sakaguchi2021WinoGrande}. To ensure evaluating fairness and reproducibility, all evaluation metrics reported in our study are measured in the same environment. It is worth mentioning that we used the raw accuracy metrics, rather than the normalized ones. The evaluation results are presented in Table~\ref{tab: commonsense}, all models are evaluated in a zero-shot setting on these tasks. We notice \llm outperforms several baseline models, particularly surpassing TinyLlama in various evaluation metrics, and it's comparable to Qwen1.5 in terms of overall performance.

\begin{table}[ht]
  \centering
  \setlength{\tabcolsep}{3pt}
  \begin{tabular}{lcccccccccc}
    \toprule
    Model & \textbf{ARC-c} & \textbf{ARC-e} & \textbf{Boolq} & \textbf{HS.} & \textbf{OB.} & \textbf{PiQA} & \textbf{SciQ} & \textbf{TQ.} & \textbf{Wino.} & \textbf{Avg}\\
    \midrule
    \rowcolor{darkgreen!20}
    OPT-1.3B\xspace&23.29&57.03&57.80&41.52&23.20&71.71&84.30&7.48&59.59&47.32 \\
    \rowcolor{darkgreen!20}
    Pythia-1.4B\xspace&25.60&57.58&60.34&39.81&20.20&71.06&85.20&5.01&56.20&47.00 \\
    % MPT-1.3B\xspace&26.88&60.73&51.8&41.46&23.8&71.38&85.7&11.78&57.3&47.87 \\
    \rowcolor{darkgreen!20}
    TinyLLaMA-3T-1.1B\xspace&27.82&60.31&57.83&44.98&21.80&73.34&88.90&11.30&59.12&48.59  \\
    \rowcolor{darkgreen!20}
    MobileLLaMA-1.4B\xspace&26.28&61.32&57.92&42.87&23.60&71.33&87.40&12.02&58.25&49.00 \\
    \rowcolor{darkgreen!20}
    Qwen1.5-1.8B\xspace&32.25&64.69&66.48&45.49&23.80&73.45&92.90&1.01&61.17&51.25 \\
    % MobileLLaMA-2.7B-Base\xspace&0.3157&0.6675&0.6367&0.4836&0.272&0.7470&0.920&0.2811&0.6148&0.5487 \\
    % StableLM-2-1.6B\xspace&36.52&66.71&80.09&53.28&26.6&74.86&88.0&7.85&64.25&55.35 \\
    \rowcolor{darkred!20}
    H2O-danube-1.8B\xspace&32.94&67.42&65.75&50.85&27.40&75.73&91.50&25.05&62.35&55.44 \\
    \rowcolor{darkred!20}
    InternLM2-1.8B\xspace&37.54&70.20&69.48&46.52&24.40&75.57&93.90&36.67&65.67&57.77 \\
    % Phi-1.5-1.3B\xspace&44.71&76.14&74.98&47.95&38.6&76.55&93.3&7.96&72.93&59.24 \\
    % Gemma-2B\xspace&40.19&74.24&69.45&52.71&30.20&76.99&94.40&33.23&64.8&59.58 \\
    \midrule
    \rowcolor{xdcolor!20}
    \llm-1.1B\xspace&28.16&62.29&61.44&45.96&24.00&72.03&89.70&18.46&60.62&51.41   \\
    \bottomrule
  \end{tabular}
  \newline
  \caption{Performance on commonsense reasoning tasks. Models marked in green perform worse than \llm, while models marked in red perform better than \llm.}
  \label{tab: commonsense}
\end{table}

\noindent\textbf{Problem-solving evaluation} For exploring the performance of the model beyond common-sense reasoning, we also evaluate the model's problem-solving capability. Specifically, our evaluation comprised: 
\begin{itemize}
\item BIG-Bench Hard (BBH) \citep{Suzgun2022ChallengingBT}: this is a subset of 23 challenging tasks from the BIG-Bench benchmark \citep{srivastava2023imitation} designed to gauge the proficiency of a language model in comprehending and executing complex instructions.
\item The General Language Understanding Evaluation (GLUE) \citep{wang-etal-2018-glue}: this is a collection of resources for training, evaluating, and analyzing natural language understanding systems.
\item Grade School Math 8K (GSM8k) \citep{cobbe2021gsm8k}: this comprises 8.5k high-quality grade school math word problems, diverse in linguistic aspects. It was curated specifically for question answering tasks involving multi-step reasoning on basic mathematical problems.
\item Massive Multitask Language Understanding (MMLU) \citep{hendryckstest2021}: this is used to measure a model's breadth of world knowledge and its problem-solving proficiencies across diverse subjects. 
\end{itemize}

The evaluation results are presented in Table~\ref{tab: problem-solving}. It can be observed that the \llm has achieved the highest performance on BBH compared to the baseline models, and it demonstrates strong competitiveness overall.

\begin{table}[ht]
  \centering
  \setlength{\tabcolsep}{8pt}
  \begin{tabular}{lcccccc}
    \toprule
    Model & \textbf{BBH} & \textbf{GLUE} & \textbf{GSM8K} & \textbf{MMLU} & \textbf{Avg} & \textbf{Avg}\\
    & 3-shot& 5-shot& 5-shot& 5-shot&&w.o. GSM8k \\
    \midrule
    % Gemma-2B\xspace   &36.51&54.15&17.97&34.00&35.66 \\
    \rowcolor{darkgreen!20}
    OPT-1.3B\xspace   &22.67&51.06&0.83&26.70&25.32&33.48 \\
    \rowcolor{darkgreen!20}
    Pythia-1.4B\xspace   &25.37&52.23&1.63&25.40&26.16&34.33 \\
    \rowcolor{darkgreen!20}
    MobileLLaMA-1.4B\xspace   &23.48&43.34&1.44&24.60&23.22&30.47 \\
    \rowcolor{darkgreen!20}
    TinyLLaMA-3T-1.1B\xspace   &26.75&48.25&1.97&25.70&25.67&33.57 \\
    \rowcolor{darkgreen!20}
    H2O-danube-1.8B\xspace  &27.31&49.83&1.90&25.70&26.19&34.28 \\
    \rowcolor{darkred!20}
    InternLM2-1.8B\xspace  &16.86&58.96&23.50&42.00&35.34&39.27 \\
    \rowcolor{darkred!20}
    Qwen1.5-1.8B\xspace &13.84&64.57&33.59&45.10&39.28&41.17 \\
    \midrule
    \rowcolor{xdcolor!20}
    \llm-1.1B\xspace    &27.34&52.61&2.58&25.90&27.11&35.28 \\
    \bottomrule
  \end{tabular}
  \newline
  \caption{Performance on problem-solving tasks. Models marked in green perform worse than \llm, while models marked in red perform better than \llm.}
  \label{tab: problem-solving}
\end{table}

\noindent\textbf{Chinese ability} Besides evaluating the model's English proficiency, we also conducted assessments on its Chinese language capabilities due to the presence of a certain proportion of Chinese content in our corpus. Specifically, our evaluation comprised:
\begin{itemize}
%\item C-Eval \citep{huang2023ceval}, this is a comprehensive Chinese evaluation suite for foundation models. It consists of 13948 multi-choice questions spanning 52 diverse disciplines and four difficulty levels.
%\item CMMLU \citep{li2023cmmlu}, this is a comprehensive evaluation benchmark specifically designed to evaluate the knowledge and reasoning abilities of LLMs within the context of Chinese language and culture. CMMLU covers a wide range of subjects, comprising 67 topics that span from elementary to advanced professional levels.
\item ARC \citep{Clark2018ThinkYH}, this consists of 7,787 science exam questions drawn from a variety of sources, including science questions provided under license by a research partner affiliated with AI2. We utilized the Chinese translation version of the original work in our study.
\item XCOPA \citep{ponti2020xcopa}, this is designed to assess how well machine learning models can transfer commonsense reasoning across different languages. It is a translated and reannotated version of the English COPA \citep{Gordon2011ChoiceOP} and includes 11 languages from 11 different language families and various regions worldwide. 
\item XNLI \citep{conneau2018xnli}, this is designed to evaluate the ability of natural language processing systems to understand and transfer knowledge across multiple languages.
\end{itemize}

The evaluation results are presented in Table~\ref{tab: chinese}, all models are evaluated in a zero-shot setting on these tasks. We observed that by adding 15\% Chinese tokens, our model gained a certain degree of understanding and generation capabilities in Chinese, surpassing some existing models, but still weaker compared to InternLM2 and Qwen1.5.

\begin{table}[ht]
  \centering
  \setlength{\tabcolsep}{15pt}
  \begin{tabular}{lcccc}
    \toprule
    Model & \textbf{ARC-zh} & \textbf{XCOPA-zh} & \textbf{XNLI-zh}& \textbf{Avg}\\
    \midrule
    % Gemma-2B\xspace   &29.06&61.20&35.10&41.79 \\
    \rowcolor{darkgreen!20}
    OPT-1.3B\xspace   &18.80&53.00&33.45&35.08 \\
    \rowcolor{darkgreen!20}
    Pythia-1.4B\xspace   &21.03&52.60&34.06&35.90 \\
    \rowcolor{darkgreen!20}
    MobileLLaMA-1.4B\xspace   &20.26&52.80&33.82&35.63 \\
    \rowcolor{darkgreen!20}
    TinyLLaMA-3T-1.1B\xspace   &21.37&56.80&33.25&37.14 \\
    \rowcolor{darkgreen!20}
    H2O-danube-1.8B\xspace  &21.79&55.60&34.74&37.38 \\
    \rowcolor{darkred!20}
    InternLM2-1.8B\xspace   &27.69&66.80&34.58&43.00 \\
    \rowcolor{darkred!20}
    Qwen1.5-1.8B\xspace &32.14&66.00&39.28&45.81 \\
    \midrule
    \rowcolor{xdcolor!20}
    \llm-1.1B\xspace    &26.24&60.60&36.02&40.95 \\
    \bottomrule
  \end{tabular}
  \newline
  \caption{Performance on Chinese tasks. Models marked in green perform worse than \llm, while models marked in red perform better than \llm.}
  \label{tab: chinese}
\end{table}

\section{Case study}
\label{sec:case study}

\subsection{Evolution of model's performance}

\begin{figure}[ht]
\centering
\includegraphics[width=\linewidth]{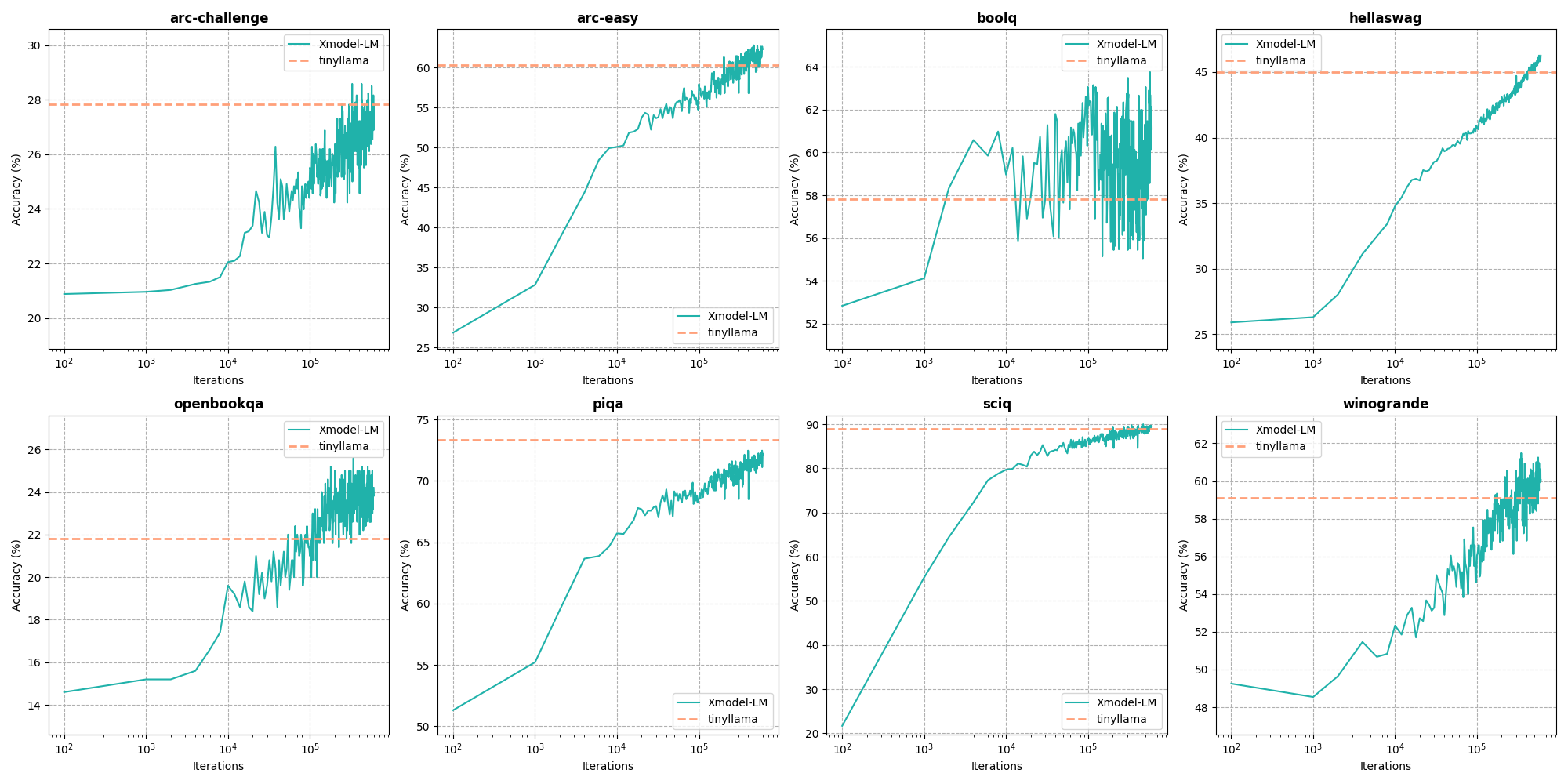}
\caption{Evolution of performance in commonsense reasoning tasks during pre-training}
\label{fig:evolution}
\end{figure}

We tracked and recorded the model's performance on the common-sense reasoning tasks during the pretraining process, as shown in Figure~\ref{fig:evolution}. It can be observed that the performance of \llm improves as training progresses, surpassing TinyLLaMA in multiple tasks. Moreover, we observe an approximate linear link between log iteration steps and model metrics gains across most tasks.

\subsection{Generalization vs memorization}

\begin{figure}[ht]
\centering
\includegraphics[width=340pt]{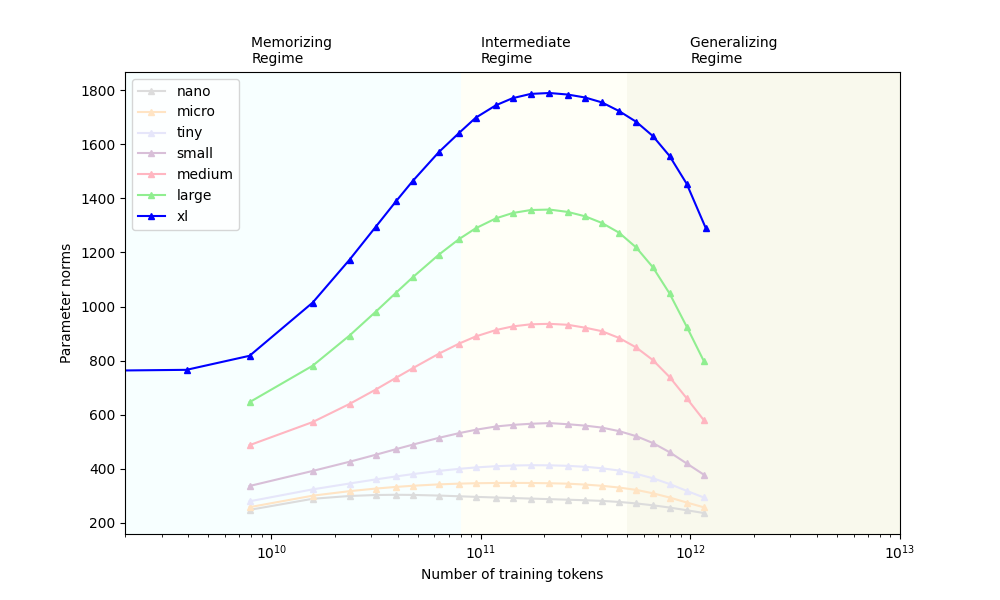}
\caption{Shifts in the \(L_2\)-norm of parameters during pre-training}
\label{fig:norm-trend}
\end{figure}

Recent advancements in mechanical interpretability have uncovered a phenomenon known as superposition, which can represent more features than expected in neural networks. During the pre-training of \llm, we observed shifts in the \(L_2\)-norm (\(\| \theta \|_2\)) of the model parameters, mirroring trends identified in prior research \citep{Tom2023Super}. Specifically, the training process can be roughly delineated into three stages, as depicted in Figure~\ref{fig:norm-trend}. 

This replication underscores a similar tripartite division, the model is in a memorization state when the amount of training data is small, during which the \(L_2\)-norm of the model gradually increases. After passing through an intermediate state, the model transitions to a generalization state, during which the \(L_2\)-norm of the model gradually decreases.

\section{Conclusions}
In summary, our work involves training a 1.1B-sized language model (\llm) on a self-built tokenizer and corpus containing both Chinese and English data. Our model demonstrates competitive performance on evaluation datasets compared to models of similar scales. Our efforts contribute to the advancement of knowledge by showcasing the potential of training smaller models with extensive datasets for future applications.

%\newpage

\small
\bibliography{xmodel}   % Specifies the .bib file (without the extension)
\end{document}